\documentclass[letterpaper, 10 pt, conference]{ieeeconf}  

\IEEEoverridecommandlockouts                              

\usepackage{graphicx} 
\usepackage{amsmath} 
\usepackage{amssymb}  
\usepackage{gensymb}
\let\labelindent\relax  
\usepackage{enumitem}

\usepackage{booktabs}
\usepackage{multirow}
\usepackage[table]{xcolor}
\usepackage[
    bookmarks=true,
    colorlinks=true,  
    linkcolor=purple,    
    citecolor=blue,  
    filecolor=magenta, 
    urlcolor=purple,     
    anchorcolor=blue
]{hyperref}
\usepackage{cite}

\title{\LARGE \bf
MISCGrasp: Leveraging Multiple Integrated Scales and Contrastive Learning for Enhanced Volumetric Grasping
}

\author{Qingyu Fan$^{1,2}$, Yinghao Cai$^{1\dagger}$, Chao Li$^{3}$, Chunting Jiao$^{3}$, Xudong Zheng$^{3}$, Tao Lu$^{1}$,  Bin Liang$^{3}$, Shuo Wang$^{1, 2}$
\thanks{$^{1}$State Key Laboratory of Multimodal Artificial Intelligence Systems, Institute of Automation, Chinese Academy of Sciences.}%
\thanks{$^{2}$School of Artificial Intelligence, University of Chinese Academy of Sciences.}%
\thanks{$^{3}$Qiyuan Lab.}%
\thanks{*This work was supported in part by the National Natural Science Foundation of China under Grants 62273342, 62473366 and U23B2038, and in part by the Qiyuan Lab Innovation Fund Project (2022-JCJQ-LA-001-024).}
\thanks{$\dagger$Corresponding to {\tt \small yinghao.cai@ia.ac.cn}.}%
}

\begin{document}

\maketitle
\thispagestyle{empty}
\pagestyle{empty}

\begin{abstract}


Robotic grasping faces challenges in adapting to objects with varying shapes and sizes. In this paper, we introduce MISCGrasp, a volumetric grasping method that integrates multi-scale feature extraction with contrastive feature enhancement for self-adaptive grasping. We propose a query-based interaction between high-level and low-level features through the Insight Transformer, while the Empower Transformer selectively attends to the highest-level features, which synergistically strikes a balance between focusing on fine geometric details and overall geometric structures. Furthermore, MISCGrasp utilizes multi-scale contrastive learning to exploit similarities among positive grasp samples, ensuring consistency across multi-scale features. Extensive experiments in both simulated and real-world environments demonstrate that MISCGrasp outperforms baseline and variant methods in tabletop decluttering tasks. More details are available at \url{https://miscgrasp.github.io/}.

\end{abstract}

\section{INTRODUCTION}
In recent years, significant advancements have been made in visual perception-based 6-DoF robotic grasping  \cite{fang2020graspnet,sundermeyer2021contact,wang2021graspness,breyer2021volumetric,jiang2021synergies,hu2024orbitgrasp}. However, robotic grasping remains an open challenge due to the vast variety of object shapes and sizes in real-world environments. The ability to predict reliable 6-DoF grasp poses that ensures stability while accommodating diverse object geometries is essential. According to \cite{zhang1994robustness,feix2015grasp,saudabayev2018human,park2021softness,pan2021emergent}, two-finger parallel grasp can be categorized into two types: 
\begin{itemize}
    \item \textbf{Power Grasp}: Achieves force closure through gripping the main portion or the entire surface of an object, with the gripper covering and stabilizing the core structure to ensure a secure hold. This grasp configuration is suitable for objects with regular shapes and sizes below gripper clearance.
    \item \textbf{Pinch Grasp}: Implements form closure by gripping the edges, protrusions, or extended parts of an object. This configuration is more suitable for objects with irregular shapes, which allows for stable grasping across different shapes and sizes through precise grasp point selection.
\end{itemize}

\begin{figure}[t]
    \centering
    \includegraphics[width=0.49\textwidth]{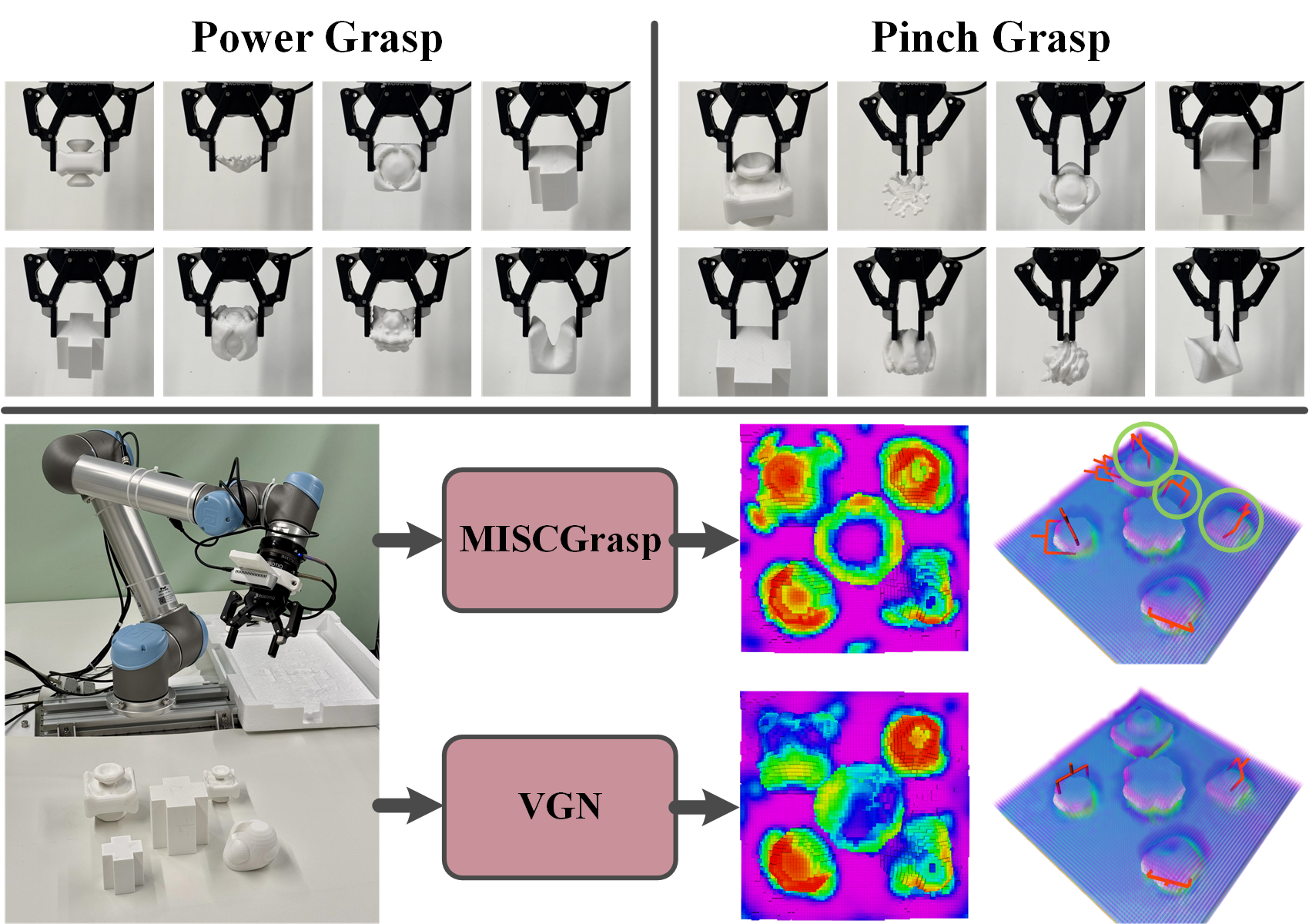}
    \caption{\textbf{Overview of MISCGrasp.} 
    The upper part showcases examples of power and pinch grasps. The lower part shows the grasp qualities and poses predicted by our method and VGN \cite{breyer2021volumetric}. The green circles highlight the differences in predictions. Our method gives different grasp predictions for objects of the same shape but different sizes.}
    \vspace{-0.5cm}
    \label{fig1}
\end{figure}

Fig. \ref{fig1} showcases examples of power and pinch grasps. Most existing works  \cite{fang2020graspnet,sundermeyer2021contact,wang2021graspness,breyer2021volumetric,jiang2021synergies,hu2024orbitgrasp} primarily focus on power grasps, where the objects can often be fitted within the gripper's width. However, real-world scenarios require both power and pinch grasps. 
Some objects are suitable for power grasps, while others are too large and can only be grasped using a pinch grasp. In this paper, we introduce \textbf{MISCGrasp}, a grasping framework that achieves self-adaptive 6-DoF manipulation through synergistic integration of power and pinch grasping paradigms.

To effectively evaluate the performance of both power and pinch grasping, a dataset containing a diverse range of objects with complex geometric features is necessary. However, most existing datasets \cite{breyer2021volumetric, fang2020graspnet, eppner2021acronym} fall short in terms of diversity, primarily featuring simple or regularly shaped objects. These limitations hinder the evaluation of 6-DoF grasping capable of handling a wide variety of object sizes and shapes, especially those requiring pinch grasps. The Evolved Grasping Analysis Dataset (EGAD) \cite{morrison2020egad} utilizes evolutionary algorithms to generate objects with diverse forms, ranging from basic geometric primitives to intricate, abstract, or organic structures with multiple protrusions and cavities. Therefore, we use EGAD as the object set and apply the data generation pipeline from \cite{breyer2021volumetric} to create a dataset rich in both power and pinch grasps.


The key challenge in handling a wide variety of objects lies in the difficulty of detecting geometric features across different scales. As convolutional layers deepen, the fine-grained geometric details necessary for pinch grasp detection tend to diminish due to feature abstraction. To this end, we propose a multi-scale feature integration approach under sparse supervision \cite{breyer2021volumetric,wu2024economic}, which selectively captures and fuses both low-level detailed features and high-level contextual features. 

Additionally, to further exploit the potential while ensuring consistency among multi-scale geometric features, we propose a self-supervised feature enhancement using contrastive learning, which fully leverages the similarities among positive samples across the entire dataset to help the model better characterize and align the underlying explicit representations for grasping.

We conduct a comprehensive evaluation of MISCGrasp in both simulated and physical environments. Compared to the original implementation of VGN \cite{breyer2021volumetric}, our approach improves the declutter rate by $25.5\%$ in experiments with an emphasis on pinch grasping. Ablation experiments are also carried out to evaluate the importance of each component of our proposed method. In summary, our contributions are:
\begin{itemize}[itemsep=0pt, topsep=0pt]
    \item We introduce MISCGrasp, a framework that enables self-adaptive 6-DoF grasping by seamlessly integrating power and pinch grasps. By incorporating geometric features at multiple scales, MISCGrasp improves the ability of grasping to handle a wide variety of objects.
    \item A self-supervised feature enhancement method is proposed which further exploits the potential of multi-scale geometric features and ensures consistency among them at  each level using contrastive learning.
    \item We generate a grasping dataset rich in both power and pinch grasps using a geometrically diverse object set, which lays the foundation for  evaluating 6-DoF grasping of objects of different shapes and sizes. Through extensive experiments, we demonstrate that MISCGrasp significantly outperforms baseline methods and variants, especially in scenarios suitable for pinch grasp.
\end{itemize}

\section{RELATED WORK}

\subsection{Grasping Dataset}

BigBird \cite{singh2014bigbird} and YCB \cite{calli2015ycb} provide high-resolution 3D object scans for robotic grasping. However, the object diversity is limited primarily to  household items. Virtual datasets, derived from 3D mesh repositories \cite{shilane2004princeton, mahler2016dex, wohlkinger20123dnet, kasper2012kit}, are primarily designed for 3D recognition task and often lack the semantic and geometric diversity necessary for robotic grasping. Tobin et al. \cite{tobin2018domain} proposed to generate simulated objects using convex primitives. EGAD \cite{morrison2020egad} further increases the diversity of object geometries with evolutionary algorithms, which is well-suited for evaluating pinch grasps.

The Cornell Grasp Dataset \cite{jiang2011efficient} provides annotated rectangle-based grasp labels for parallel grasp detection. Dex-Net \cite{mahler2016dex} extended the dataset by synthesizing millions of grasp samples. GraspNet-1B dataset \cite{fang2020graspnet} contains $97280$ RGB-D images with over one billion grasp poses. These datasets have abundant grasp labels for supervision which is called dense supervision in \cite{wu2024economic}. However, obtaining dense grasp annotations is resource-intensive, which also makes the training process difficult to converge. VGN \cite{breyer2021volumetric} introduced a resource-efficient  pipeline for simulating grasp attempts in cluttered scenes, where grasp labels are obtained through grasping trials. In this paper, we adopt the grasp data generation pipeline from \cite{breyer2021volumetric} with several modifications to improve the efficiency of the grasp sampling process.


\subsection{Deep Robotic Grasping}

Recent advances in robotic grasping mainly use deep learning to predict grasp poses from sensory data \cite{fang2020graspnet, sundermeyer2021contact, wang2021graspness}. While these methods have achieved impressive grasping performance, they generally use dense supervision, which require huge resource costs for model training. In contrast, sparse supervision—using only a few grasp samples per object or scene—has been explored in
\cite{wu2024economic,breyer2021volumetric,jiang2021synergies}. Sparse supervision generally has limited grasp performance due to inadequate supervision. How to utilize sparse supervision for effective grasping remains unsolved. VGN \cite{breyer2021volumetric} is a volumetric grasping network that learns collision-free grasps in parallel. GIGA \cite{jiang2021synergies} combines grasp prediction with geometry reconstruction. The object diversities are limited to learn both power and pinch grasps. The fine-grained geometric details necessary for pinch grasp detection
tend to diminish with increasing network depth.
In this paper, we propose a multi-scale feature integration approach that enables self-adaptive grasping by balancing attention between small graspable areas and overall geometric structures.

\subsection{Contrastive Learning in Grasping}

Contrastive learning \cite{hadsell2006dimensionality} aims to bring positive pairs closer and push negative pairs apart. SimCLR \cite{chen2020simple} relies on large batch sizes for negative samples, while MoCo \cite{he2020momentum} uses a negative sample queue and a momentum encoder for consistency. SimSiam \cite{chen2021exploring} avoids collapse with a stop-gradient operation, enabling learning without negatives. NNCLR \cite{dwibedi2021little} uses a nearest-neighbor memory bank to reinforce positive pair proximity.


In the context of grasping, Zhu et al. \cite{zhu20216} applies contrastive learning to depth features for sim-to-real transfer. Liu et al. \cite{liu2023self} enhances affordance cues with a Siamese encoder and triplet loss. Wang et al. \cite{wang2024graspcontrast} uses self-supervised pre-training to improve 6-DoF grasp detection. In contrast, our approach applies contrastive learning to multi-scale features of positive grasp samples, with an aim to improve the feature representation while maintaining feature consistency across different scales.

\begin{figure*}[t]
    \centering
    \includegraphics[width=\textwidth]{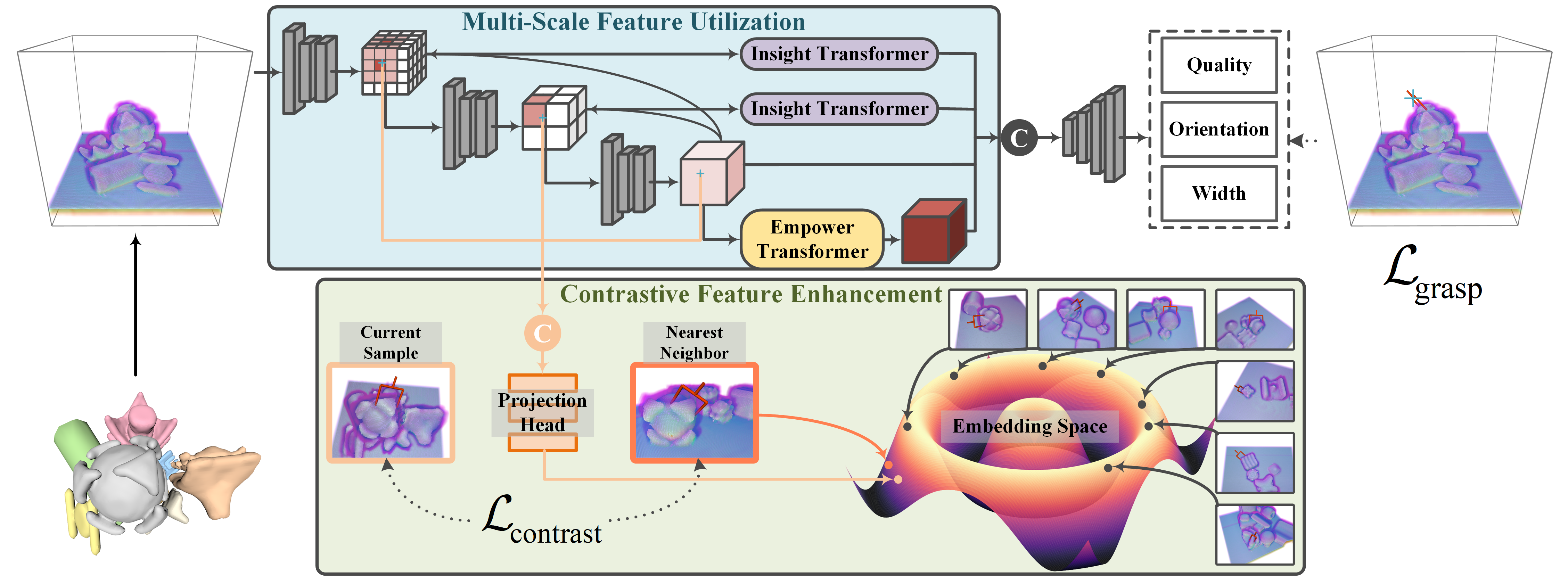}
    \caption{\textbf{Framework of MISCGrasp.} MISCGrasp utilizes multi-scale features and contrastive feature enhancement for self-adaptive 6-DoF grasping. The Multi-scale Feature Utilization Module extracts and integrates diverse geometric features from different scales. The Insight Transformer facilitates query-based interactions, enriching high-level features with fine-grained details from lower-level features, while the Empower Transformer selectively attends to self-information from the highest-level features. After processing through these transformers, the multi-scale features are concatenated and decoded to produce grasp predictions. The Contrastive Feature Enhancement Module refines the feature representations by pulling multi-scale features from similar positive grasp samples closer together in the latent space. This module reinforces the relevance of features and ensures consistency across different scales.}
    \label{fig2}
    \vspace{-0.5cm}
\end{figure*}

\section{METHOD}

\subsection{Problem Formulation}

The 6-DoF grasp detection task involves identifying feasible grasp poses within  $SE\left(3\right)$. Specifically, the task of volumetric grasping is to predict grasp configurations given a Truncated Signed Distance Function (TSDF) fused by multiple depth images:$$
F:\boldsymbol{\Omega} \to \left\{ {{q_i},{\mathbf{r}_i},{w_i}} \right\}_{i = 1}^{{N^3}} 
$$where the TSDF $\boldsymbol{\Omega}$ is discretized into an \( N^3 \) voxel grid. The grasp representation follows the definition in \cite{breyer2021volumetric}, where the grasp quality \( q \in [0, 1] \), the orientation \( \mathbf{r} \in SO(3) \), and the grasp width \( w \in \mathbb{R} \). The framework of MISCGrasp is illustrated in Fig. \ref{fig2}.

\subsection{Grasp Data Generation}

Most existing grasping datasets \cite{fang2020graspnet}, \cite{sundermeyer2021contact}, \cite{breyer2021volumetric} lack sufficient diversity in terms of object shapes and sizes. Data-driven methods can be easily biased by the distribution of the data. As a result, the grasping models training from these datasets tend to learn power grasps. The objects in EGAD  \cite{morrison2020egad} are generated using evolutionary algorithms, ensuring a wide range of geometry complexity and grasp difficulty. This diversity in object shapes provides basis for learning pinch grasps, which is why EGAD objects are introduced as the target object set in our work.

We adopt the grasp data generation pipeline from \cite{breyer2021volumetric} with several modifications. After creating the scene in simulation, a point cloud is reconstructed from 6 depth images captured from multiple views, and a grasp candidate from the surface points is sampled with a random offset along the normal. Six grasp orientations are evaluated which are evenly spaced along a semi-circle to account for the gripper's symmetry. The pipeline generates sparse ground-truth grasp labels through grasping trials.


Different from \cite{breyer2021volumetric}, where objects are resized to fit the gripper, we apply a random scaling $s \in \mathcal{U}\left( 0.65,1.7 \right)$ relative to the gripper's opening width to balance the collection of both power and pinch grasp data. It is noted that the most time-consuming step in the original grasp generation pipeline is the grasping trials. To improve efficiency, we design a pre-pruning method. The parallel gripper is simplified to a five-keypoint model. The positions of the gripper keypoints are computed from the sampled pre-grasp and grasp poses. This allows us to use TSDF to check for collisions instead of performing full grasp attempts. Invalid grasp poses are pruned during this step, which reduces the search space and improves the efficiency of the grasp pose sampling process.

\subsection{Multi-Scale Feature Utilization}

Due to the regularity of volumetric grids, 3D convolutional neural networks are well-suited for per-voxel grasp prediction. The coarse-to-fine multi-scale features are learned through stacked convolutional layers. However, high-level CNN features often focus on global information, neglecting crucial fine-grained geometric details. This may lead to missing pinch grasps, especially for small objects or specific graspable areas. To utilize multi-scale features, we employ a lightweight Feature Pyramid Network (FPN) \cite{lin2017feature} as the backbone for feature extraction.

However, directly aggregating multi-scale features through skip connections can lead to inconsistencies between features at different scales, which degrade the grasping performance, as demonstrated in our experimental results. To this end, we propose the Insight Transformer (IT), inspired from \cite{woo2018cbam}, which facilitates query-based interaction from high-level to low-level features. Additionally, we propose the Empower Transformer (ET), which selectively attends to self-information from the highest-level features. Together, these methods synergistically enhance grasping performance under sparse grasp supervision.

\textbf{Insight Transformer (IT)}: Low-level features in multi-scale representations generally include detailed geometric information but lack global contextual awareness. An intuitive way of utilizing multi-scale features is to leverage high-level features to query low-level ones, which supplements the detailed fine-grained features that diminish as the layer depth increases. IT operates in a bottom-up manner, utilizing channel-wise attention. The process is expressed as follows:\[
\mathbf{A}_{\text{c}} = \text{MLP}\left( \text{AvgPool}\left( \mathbf{K} \right) \right),
\]
\[
\mathbf{X}_{\text{I}} = F_{\text{refine}}\left( \text{Conv}\left( \mathbf{A}_{\text{c}} \odot \text{Conv}\left( \mathbf{Q} \right) \right) + \text{Conv}_{\text{stride}}\left( \mathbf{V} \right) \right).
\]where the high-level feature volume is denoted as $\mathbf{Q}$, the low-level feature volume is denoted as $\mathbf{K}$ and $\mathbf{V}$. $\mathbf{K}$ first undergoes a global average pooling operation, producing the channel attention map $\mathbf{A}_{\text{c}}$ through an MLP. Next, after a smoothing convolution, $\mathbf{Q}$ is modulated by $\mathbf{A}_{\text{c}}$ and added to the downsampled $\mathbf{V}$. To align the resolution of $\mathbf{V}$ with that of the high-level feature volume, $\mathbf{V}$ is reshaped using several convolutional layers with  strides. Finally, the combined result is passed through an output layer for feature refinement.

\textbf{Empower Transformer (ET)}: Directly applying self-attention to feature volumes increases computational costs and risks of overfitting to noise. ET incorporates constraints into the self-attention mechanism and leverages the Mixture of Softmax (MoS) \cite{yang2018breaking} for self-enhancement. 

Specifically,  \(1 \times 1\) convolutions are applied to the highest-level feature volume to generate  \(\mathbf{Q}\), \(\mathbf{K}\), and \(\mathbf{V}\), ensuring that \(\mathbf{Q}\) and \(\mathbf{K}\) share the same linear transformation. However, such design may lead to overemphasis on self-attention and feature redundancy. To mitigate this, we adopt the MoS approach, which enables the model to selectively attends to the most relevant features. 

We first apply spatial average pooling to query vector to obtain the averaged query vector \(\bar{\mathbf{Q}}\). Next, the query and key vectors are divided into \(\mathcal{N}\) parts. The averaged query vector \(\bar{\mathbf{Q}}\) is multiplied with a learnable parameter matrix \(\mathbf{W} \in \mathbb{R}^{\mathcal{N} \times d_{\text{q}}}\), and normalized via softmax to produce the mixture coefficient vector \(\boldsymbol{\pi}\). The mixture coefficients are then multiplied by their corresponding attention weights and summed to generate the final attention map $\mathbf{A}_{\text{s}}$. This attention map is multiplied with the value vector, refined through a convolutional layer, and finally added to the input highest-level features $\mathbf{X}_{high}$ via a residual connection:
\[
\bar{\mathbf{Q}} = \text{AvgPool}\left( \mathbf{Q} \right), \boldsymbol{\pi}  = \text{softmax} \left( {\mathbf{W} \bar{\mathbf{Q}}} \right),
\]
\[
\mathbf{A}_{\text{s}} = \sum\limits_{n = 1}^\mathcal{N} {{\pi _n} \text{softmax} \left( {\frac{{{\mathbf{Q}^{\left( n \right)}}^\top \mathbf{K}^{\left( n \right)}}}{{\sqrt {{d_{\text{q}}}} }}} \right)},
\]
\[
\mathbf{X}_{\text{E}} = \text{Conv}\left(\mathbf{A}_{\text{s}} \mathbf{V}\right) + \mathbf{X}_{\text{high}}.
\]


Finally, we concatenate the outputs of IT, ET, and the highest-level feature, passing them through a convolutional layer for fusion before feeding the integrated multi-scale features into the decoder for grasp prediction. Experimental results demonstrate that ET contributes to performance improvement.


\subsection{Contrastive Feature Enhancement}

Contrastive learning aims to learn feature embeddings where positive sample pairs are close
to each other, while negative pairs are far apart. However, 
6-DoF grasp poses, represented in the SE(3) space, have a high-dimensional and continuous search space, making the negative samples computationally complex to sample. Therefore, we instead focus on fully exploiting the similarities among positive samples to enhance the feature representations for grasp
detection. While these similarities may not always appear within the same scene, they are abundant across the entire dataset.

In our approach, the nearest neighbor (NN) memory bank \cite{dwibedi2021little} is used to store positive embeddings. The memory bank is large enough to approximate the full dataset distribution in the embedding space. Specifically, in each training iteration, trilinear interpolation is performed on the positions of positive samples in the scene to obtain their corresponding embeddings. At the end of each training step, the memory bank is updated by concatenating the positive embeddings  of the current step and discarding the oldest ones. Meanwhile, the NN embeddings for each positive sample are obtained, and the cosine similarity loss is applied to supervise the backbone.

However, since our approach uses multi-scale feature representations, applying contrastive enhancement solely on the highest-level features may introduce feature inconsistencies across scales. Therefore, we further introduce multi-scale contrastive enhancement, which simultaneously enhances multi-scale consistency while utilizing embedding similarity, as shown in Fig \ref{fig2}. 

For volume feature $\mathbf{X}^l$ at level $l$, we interpolate the positive sample position $\mathbf{p}$ to obtain positive embeddings at different scales $\{\mathbf{x}_\mathbf{p}^1, \mathbf{x}_\mathbf{p}^2, \dots, \mathbf{x}_\mathbf{p}^L \}$. These embeddings are then concatenated before feeding to a three-layer MLP projection head $\mathcal{F}_{\text{proj}}$. The projection head aggregates multi-scale features and projects the embeddings of each positive sample into the same latent space:\[
\mathbf{x}_\mathbf{p}^l = \text{Interp}\left( \mathbf{X}^l, \mathbf{p} \right), \quad l = 1, 2, \dots, L,
\]
\[
\mathbf{z}_\mathbf{p} =  \mathcal{F}_{\text{Proj}}\left( \text{Concat}\left( \mathbf{x}_\mathbf{p}^1, \mathbf{x}_\mathbf{p}^2, \dots, \mathbf{x}_\mathbf{p}^L \right) \right),
\]
\[
\mathcal{L}_{\text{contrast}} =  - \frac{1}{|\mathbf{P}|} \sum_{\mathbf{p} \in \mathbf{P}} \frac{\mathbf{z}_{\mathbf{p}} \cdot \text{NN}(\mathbf{z}_{\mathbf{p}})}{\|\mathbf{z}_{\mathbf{p}}\| \|\text{NN}(\mathbf{z}_{\mathbf{p}})\|}.
\]
where $\text{NN}\left( \cdot \right)$ denotes the operation of retrieving the nearest neighbor from the memory bank, and $\mathbf{P}$ is the set of positive samples in the scene.

\subsection{Implementation Details}
\subsubsection{Parameter Setting}




The robot workspace is $0.4\text{m}$, which is slightly larger than in \cite{breyer2021volumetric}. Since the original spatial resolution in \cite{breyer2021volumetric} lacks sufficient geometric detail, we increase the TSDF grid resolution from $40$ to $80$.

We generate $5100$ scenes for each scene type using the EGAD and VGN packed object sets, with each scene containing $8$ positive and $64$ negative grasp labels. Of these, $5000$ scenes are used for training, and $100$ scenes for validation.

For feature extraction, we utilize an FPN with 3-level multi-scale feature volumes, with dimensions of $32$, $64$, and $128$. The ET is configured with $\mathcal{N} = 2$. The dimension of the hidden and output layers of the projection head is $128$. The size of the memory bank and embedding dimension are set to $32768$ and $128$, respectively.



\subsubsection{Training}

Each iteration processes $4$ scenes, each with $8$ positive and $16$ negative samples. Following \cite{breyer2021volumetric}, we supervise grasp quality, orientation, and gripper width. To mitigate sample imbalance, we employ a weighted binary cross-entropy loss with weights of $2$ for positive samples and $1$ for negative samples. The final loss is the weighted sum of the grasping and contrastive losses:
\[
\mathcal{L} = \mathcal{L}_{\text{grasp}} + 0.5 \times \mathcal{L}_{\text{contrast}}.
\]
The models are implemented in PyTorch \cite{paszke2019pytorch} and trained with the Adam optimizer \cite{kingma2014adam} for about $100$ epochs on an Nvidia RTX $4090$ GPU. A one-cycle learning rate scheduler \cite{smith2019super} is used, with an initial learning rate of $4\times10^{-5}$ and a maximum of $4\times10^{-4}$.

\begin{figure}[t]
    \centering
    \includegraphics[width=0.49\textwidth]{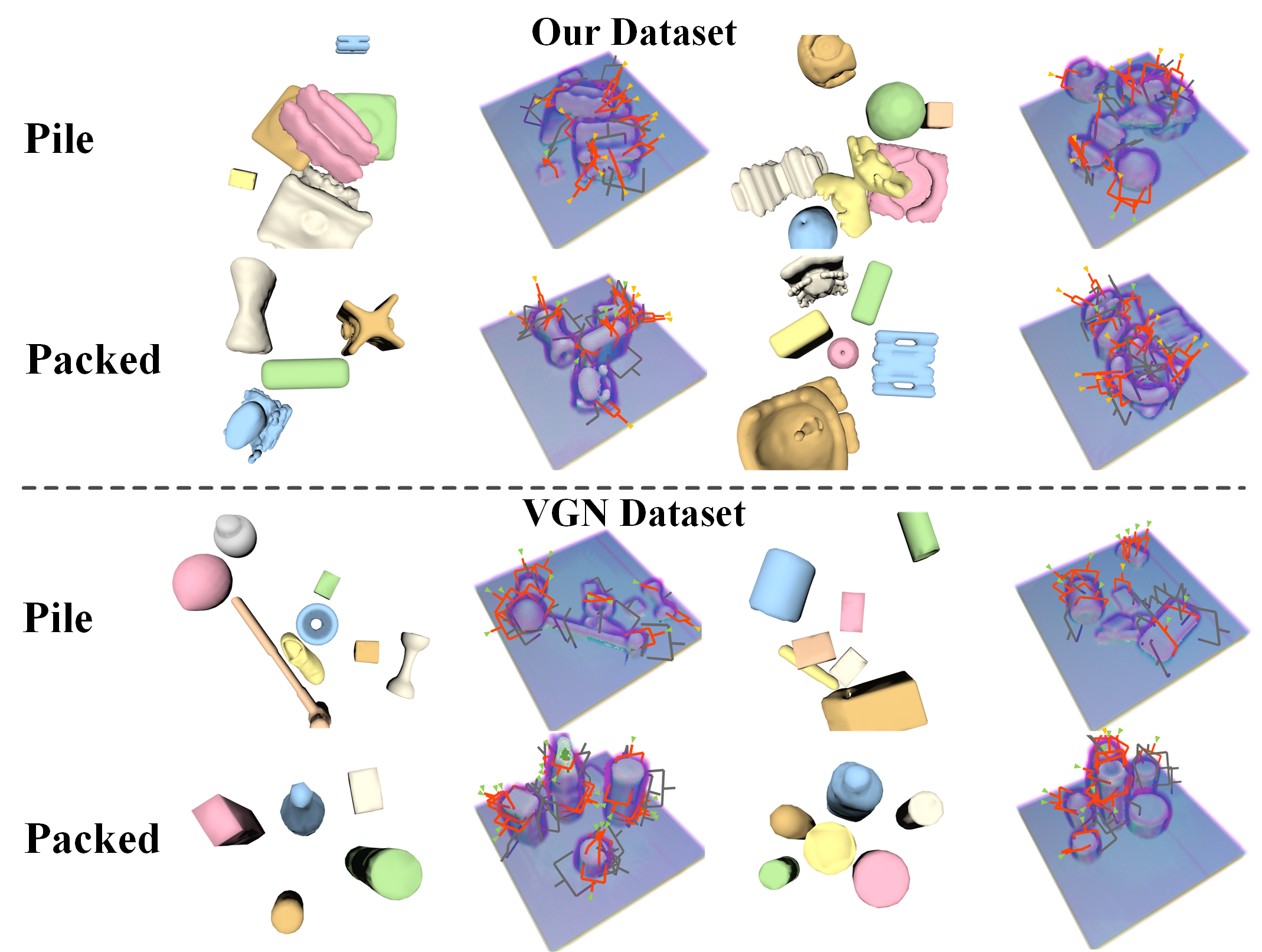}
    \vspace{-0.5cm}
    \caption{\textbf{Comparison between Our Dataset and the VGN Dataset.} Red indicates positive grasp samples, and gray indicates negative samples. The green arrows represent power grasps, and yellow arrows represent pinch grasps. Our dataset contains a diverse range of both power grasps and pinch grasps. In contrast, the VGN dataset primarily includes regularly shaped object, where pinch grasps are quite rare.}
    \label{fig3}
    \vspace{-0.5cm}
\end{figure}

\begin{table*}[t]
    \centering
        \vspace{-0.2cm}
    \caption{RESULTS OF \textbf{SINGLE-OBJECT} EXPERIMENTS IN \textbf{SIMULATION}}
    \label{tab1}
    \small
    \resizebox{0.95\textwidth}{!}{  
    \begin{tabular}{l|c|c|c|ccccccc}
        \hline
        \multirow{3}{*}{Method} & \multirow{3}{*}{Dataset} & \multirow{3}{*}{Contrastive Enhancement} & \multirow{3}{*}{Multi-scale Utilization} & \multicolumn{7}{c}{EGAD-Single} \\
        \cline{5-11}
        & & & & \multicolumn{2}{c}{Easy} & \multicolumn{2}{c}{Medium} & \multicolumn{2}{c}{Hard} & \multicolumn{1}{c}{Overall} \\
        \cline{5-11}
        & & & & SR (\%) & DR (\%) & SR (\%) & DR (\%) & SR (\%) & DR (\%) & Score \\
        \hline
        \hline
        MISCGrasp & Ours & Ours & Ours & \cellcolor{blue!20}\textbf{97.4} & \cellcolor{blue!20}\textbf{97.4} & \cellcolor{blue!20}87.1 & \cellcolor{blue!20}\textbf{78.3} \cellcolor{blue!20} & \cellcolor{blue!20}78.1 & \cellcolor{blue!20}\underline{64.1} & \cellcolor{blue!20}\textbf{234.2} \\
        \hline
        VGN & Ours & / & / & \underline{94.9} & \underline{94.9} & 82.8 & 69.6 & 70.0 & 53.8 & 202.9 \\
        VGN & VGN & / & / & 92.3 & 92.3 & 83.3 & 65.2 & \underline{79.2} & 48.7 & 194.3 \\
        \hline
        GIGA & Ours & / & / & 88.9 & 20.5 & 48.9 & 31.9 & 51.9 & 35.9 & 87.3 \\
        \hline
        MISCGrasp & Ours & Ours w/o Multi-Scale Features & Ours & \underline{94.9} & \underline{94.9} & 82.0 & 72.5 & 78.1 & \underline{64.1} & 223.0 \\
        MISCGrasp & Ours & Ours w/ Negative Samples & Ours & 92.3 & 92.3 & 86.9 & \underline{76.8} & 78.8 & \textbf{66.7} & 224.1 \\
        MISCGrasp & Ours & Ours w/ Intra-Scene Similarity & Ours & \underline{94.9} & \underline{94.9} & 85.5 & \underline{76.8} & 75.0 & 61.5 & 225.0 \\
        MISCGrasp & Ours & / & Skip Connection & 92.3 & 92.3 & \textbf{94.7} & \underline{76.8} & 70.0 & 53.8 & 217.7 \\
        \hline
        MISCGrasp & Ours & / & Ours & \underline{94.9} & \underline{94.9} & 76.6 & 71.0 & 76.5 & \textbf{66.7} & \underline{227.4} \\
        MISCGrasp & Ours & / & Ours w/o ET & 87.2 & 87.2 & 85.2 & 75.4 & 65.6 & 53.8 & 207 \\
        MISCGrasp & Ours & / & Ours w/o IT & 89.7 & 89.7 & 89.5 & 73.9 & \textbf{80.6} & \underline{64.1} & 222.1 \\
        MISCGrasp & Ours & / & / & \underline{94.9} & \underline{94.9} & \underline{91.4} & \underline{76.8} & 74.2 & 59.0 & 219.9 \\
        \hline
        MISCGrasp & Ours & / & Ours w/ Unshared ET & \underline{94.9} & \underline{94.9} & 86.7 & 75.4 & 60.0 & 53.8 & 218.5 \\
        \hline
    \end{tabular}
    }
\end{table*}

\begin{table*}[t]
    \centering
        \vspace{-0.2cm}
    \caption{RESULTS OF \textbf{MULTI-OBJECT} EXPERIMENTS IN \textbf{SIMULATION}}
    \label{tab2}
    \small
    \resizebox{0.95\textwidth}{!}{  
    \begin{tabular}{l|c|c|c|cccccc}
        \hline
        \multirow{2}{*}{Method} & \multirow{2}{*}{Dataset} & \multirow{2}{*}{Contrastive Enhancement} & \multirow{2}{*}{Multi-scale Utilization} & \multicolumn{2}{c}{EGAD+ADV-Pile} & \multicolumn{2}{c}{Pile-Pile} & \multicolumn{2}{c}{Packed-Packed} \\
        \cline{5-10}
        & & & & SR (\%) & DR (\%) & SR (\%) & DR (\%) & SR (\%) & DR (\%) \\
        \hline
        \hline
        MISCGrasp & Ours & Ours & Ours & \cellcolor{blue!20}\underline{75.2} & \cellcolor{blue!20}\underline{62.3} & \cellcolor{blue!20}\textbf{84.8} & \cellcolor{blue!20}\textbf{68.2} & \cellcolor{blue!20}89.5 & \cellcolor{blue!20}83.2 \\
        \hline
        VGN & Ours & / & / & 66.0 & 48.7 & 76.2 & 50.6 & 86.0 & 82.6 \\
        VGN & VGN & / & / & 58.0 & 36.8 & 75.3 & 47.6 & 89.3 & 77.3 \\
        \hline
        GIGA & Ours & / & / & 14.8 & 4.6 & 22.8 & 6.4 & 20.1 & 5.4 \\
        \hline
        MISCGrasp & Ours & Ours w/o Multi-Scale Features & Ours & 71.6 & 58.2 & 79.8 & 56.0 & 92.3 & 84.6 \\
        MISCGrasp & Ours & Ours w/ Negative Samples & Ours & 74.6 & 62.1 & 84.4 & 64.6 & 91.9 & \underline{85.2} \\
        MISCGrasp & Ours & Ours w/ Intra-Scene Similarity & Ours & 73.1 & 58.9 & 84.3 & 67.0 & \textbf{93.4} & \textbf{87.7} \\
        MISCGrasp & Ours & / & Skip Connection & \textbf{76.9} & \textbf{62.5} & \underline{84.7} & 61.3 & 89.6 & 80.5 \\
        \hline
        MISCGrasp & Ours & / & Ours & 73.1 & 57.7 & 81.5 & \underline{68.0} & 89.6 & 84.6 \\
        MISCGrasp & Ours & / & Ours w/o ET & 73.5 & 60.9 & 79.6 & 62.7 & 88.9 & 82.3 \\
        MISCGrasp & Ours & / & Ours w/o IT & 69.6 & 52.8 & 82.2 & 58.4 & \underline{92.4} & 83.7 \\
        MISCGrasp & Ours & / & / & 73.8 & 55.8 & 84.6 & 58.3 & 92.0 & 84.3 \\
        \hline
        MISCGrasp & Ours & / & Ours w/ Unshared ET & 70.0 & 50.9 & 78.6 & 61.8 & 91.5 & 84.8 \\
        \hline
    \end{tabular}
    }
    \vspace{-0.5cm}
\end{table*}

\section{EXPERIMENTAL RESULTS AND ANALYSIS}

\subsection{Baseline Methods}
\subsubsection{VGN \cite{breyer2021volumetric}}  
A fully convolutional approach for volumetric grasping that generates grasps  from an input TSDF volume.

\subsubsection{GIGA \cite{jiang2021synergies}} 
An implicit neural representation based approach that achieves both grasping and reconstruction. For a fair comparison, GIGA is trained with a full TSDF volume as input.



In our comparative experiments, we evaluate our MISCGrasp alongside baseline approaches, as well as their respective variants. The network depths of the baseline approaches are adjusted to accommodate higher grid resolution as in our setting. All methods are trained with the same experimental settings, unless stated otherwise. It is worth noting that several recent approaches \cite{huang2023edge,zuru2024icgnet,hu2024orbitgrasp} achieved state-of-the-art performance on the decluttering dataset introduced in \cite{breyer2021volumetric}, which is also used in our experiments. These approaches use point clouds as input and focus on grasp representation learning while our approach and \cite{breyer2021volumetric,jiang2021synergies} use TSDF fused from multiple observations. Given the differences of the input, direct comparisons with these approaches would not be fair.


\subsection{Simulation Experiments}
\subsubsection{Setup}

We follow the experimental setting in \cite{breyer2021volumetric} for a fair comparison. The simulated environment is built in PyBullet \cite{coumans2016pybullet}. A free-floating Franka Emika gripper is employed to declutter objects within a $0.4 \times 0.4 \times 0.4 \ \text{m}^3$ workspace. The input consists of a TSDF grid with $N = 80$, which is integrated from six depth images using the implementation from Open3D \cite{zhou2018open3d}. 

Three scene types are simulated: \textbf{Single}, with a single object placed at the tabletop center; \textbf{Pile}, where multiple objects are dropped to form a cluttered pile; and \textbf{Packed}, where multiple objects are arranged upright in random positions to create a densely packed layout. Grasping performance is evaluated using two metrics: \textbf{Success Rate} (SR) and \textbf{Declutter Rate} (DR), defined as the ratio of successful grasps to total grasps and the ratio of decluttered objects to total objects, respectively.

\begin{figure}[t]
    \centering
    \includegraphics[width=0.49\textwidth]{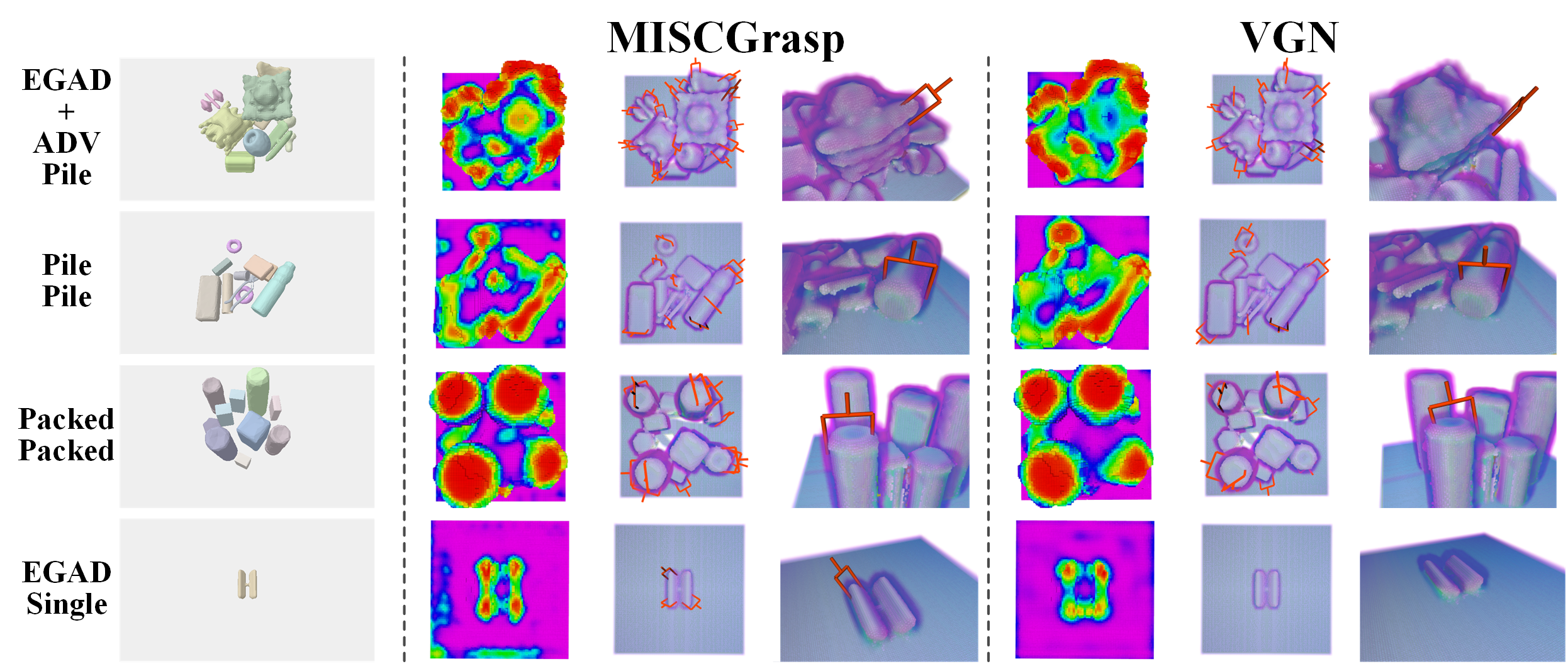}
    \vspace{-0.5cm}
    \caption{\textbf{Comparison of Visualization Results in Simulation Experiments between MISCGrasp and VGN.} The visualized grasps are filtered according to a quality score greater than $0.9$. The results demonstrate that our method predicts more diverse and feasible grasps compared to VGN. Additionally, our method produces more accurate voxel-wise grasp quality volumes.}
    \label{fig4}
    \vspace{-0.5cm}
\end{figure}

\subsubsection{Experiment Types}

We design experiments  with a focus on evaluating both pinch grasp and power grasp:


\begin{itemize}
\item \textbf{EGAD-Single}: We conduct single-scene grasping experiments using the EGAD test set. The $49$ objects in the set are scaled into $147$ objects by creating small, medium, and large variations of each object. Following \cite{morrison2020egad}, each object is scored based on its geometric complexity and grasp difficulty. These scores are then ranked and divided into three categories: easy, medium, and hard. The total scores for each category are normalized to a value of $100$. More details can be found in the supplementary materials due to the limited space.

    \item \textbf{Pile-Pile} and \textbf{Packed-Packed}:    The pile and packed test sets are used to evaluate the grasping performance in pile and packed scenes. We increase the number of objects per scene from $5$ (as used in \cite{breyer2021volumetric}) to $10$, thereby increasing the scene complexity. All other settings remain the same as in \cite{breyer2021volumetric}. This experiment is designed to evaluate the performance with regular objects.

    \item \textbf{EGAD+Adv-Pile}: The Berkeley adversarial object set \cite{mahler2019learning} includes $11$ workpiece objects and $3$ irregular objects. These, along with the EGAD test set, provide sufficient geometric complexity for evaluating pinch grasp. Using a total of $63 (11+3+49)$ objects, we generate pile scenes with random scaling of objects consistent with the training set. Each scene contains $10$ objects. This experiment is specifically designed to test the performance on pinch grasp.

\end{itemize}


We perform $147$ rounds of grasping for single-object scenarios and $100$ rounds for multi-object scenarios. After each prediction, the highest-scored grasp is executed. The gripper continues executing grasps until the workspace is cleared, a prediction fails, or the maximum number of consecutive grasp failures is reached, which is set to $1$ in the single-object scenario and $2$ in the multi-object scenario.

\subsubsection{Results}

We also regenerate the training data using the VGN object set as in \cite{breyer2021volumetric} to ensure a fair comparison. The generated VGN dataset has the same number of scenes and overall count of positive and negative grasp samples as in our dataset. We do not scale objects in VGN dataset. All other settings  remain consistent with those used in our experiments. It is observed in Tab. \ref{tab1} and Tab. \ref{tab2} that VGN trained on our dataset generally achieve higher declutter rates compared to that trained on the VGN dataset. 

Notably, in the EGAD+ADV-Pile experiment, VGN trained on our dataset outperforms that trained on the VGN dataset by a large margin. This indicates that our dataset contains diverse object geometries, enabling the model to learn both power and pinch grasps. Although there are a few cases with slightly lower success rates, we attribute this to the trade-off between predicting more diverse grasps and feasible grasping executions. Visualizations of our dataset and the VGN dataset are shown in Fig. \ref{fig3}. These visualizations highlight that our dataset is rich in pinch grasps, which are not well-covered in the VGN dataset.

While VGN performs effectively in most cases, it underperforms compared to our method in all metrics, which indicates the effectiveness of multi-scale feature utilization and contrastive enhancement. The limited performance of GIGA is likely due to the smaller dataset size used in our experiments compared to the one used in GIGA. Additionally, the TSDF grid resolution is set to $80$ in our experiments, differing from the $40$ used in GIGA. Comparisons between MISCGrasp and VGN, as shown in Fig. \ref{fig4}, demonstrate that our method generates more diverse and feasible grasps. For example, in the EGAD-Single experiment, our method successfully predicts grasps on curved and raised edges which are absent from the predictions of VGN. In the Pile-Pile experiment, VGN focuses overly on the edge of rectangular prism, while our method generates feasible grasp that spans across the object.

To validate the effectiveness of each component of our method,  we evaluate the performance of three alternatives to the contrastive enhancement module: (1) a variant using only the highest-level features for feature enhancement, (2) a variant using InfoNCE loss with both positive and negative samples, and (3) a variant that do not use the memory bank, focusing solely on intra-scene positive sample similarity. Additionally, we replace our multi-scale approach with layer-wise fusion with skip connections \cite{ronneberger2015u}.

It is observed in Tab. \ref{tab2} that the variant without multi-scale contrastive enhancement suffers significant degradation in pile experiments. This indicates that relying solely on the highest-level features lacks fine-grained  geometric features needed for pinch grasp. The variant with negative samples performs slightly worse in most metrics. We attribute this to the complex distribution of negative grasps in grasping tasks. Some negative samples are hardly separable from positives, causing noisy contrastive learning signals. The intra-scene similarity variant performs well in the Packed-Packed experiment where scenes are dominated by power grasps with clear positive-negative separability but is overall inferior to our method, particularly in the EGAD+ADV-Pile experiment. Skip connections excel in EGAD+ADV-Pile but underperform in Packed-Packed. In cluttered scenes, layer-wise fusion introduces redundant details, overwhelming contextual global awareness, which are essential for grasp detection in dense scenes. Overall, while some alternatives excel in specific experiments, our design demonstrate optimal performance across all evaluated scenarios.

\begin{figure}[t]
    \centering
    \includegraphics[width=0.49\textwidth]{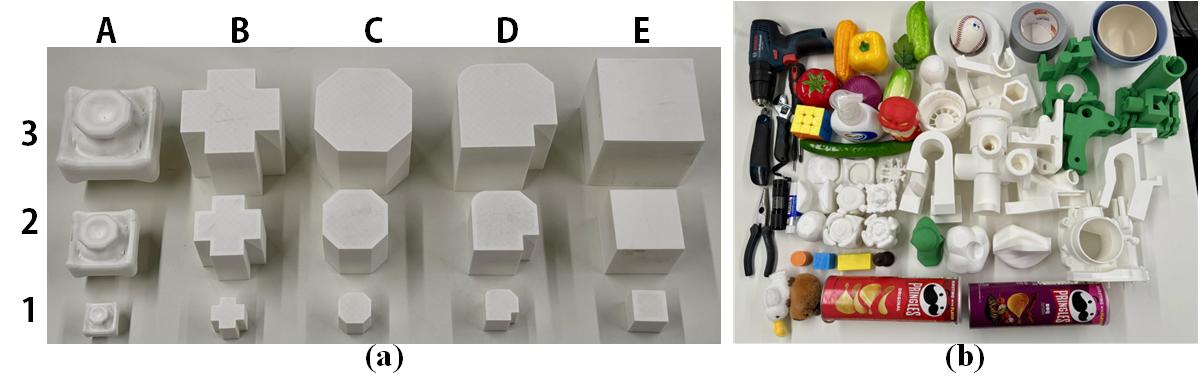}
    \vspace{-0.5cm}
    \caption{\textbf{Physical Experiment Objects.} (a) Objects used for single-object experiments. (b) Objects used for multi-object experiments.}
    \label{fig5}
    \vspace{-0.5cm}
\end{figure}

\subsubsection{Ablation Studies}

In our ablation studies, we evaluate the impact of each component in MISCGrasp:

\textbf{Contrastive Feature Enhancement.} Removing contrastive enhancement results in a performance drop especially in the easy and medium sets of the EGAD-Single experiment, though a slight improvement is observed on the difficult set. This highlights the role of contrastive enhancement across scales in balancing global and local information, which maintains  consistency across features at different levels.

There is a slight performance drop in the Packed-Packed results with contrastive enhancement. Some alternatives and ablated variants slightly outperform our full method in this setting. However, they fail to achieve consistent superiority across all experiments, particularly in pile scenes.

\textbf{Multi-Scale Feature Utilization.} Removing the ET module results in a significant performance drop in single-object experiments, which may due to the overemphasis of low-level information during feature aggregation. Ablating the IT module causes notable degradation in multi-object experiments, 
especially in declutter rates for pile scenarios, emphasizing the importance of multi-scale information in complex scenes. Removing both ET and IT leads to an overall performance decrease, suggesting their complementary roles in multi-scale feature utilization. Even with both ablated, it still outperforms the VGN, which we attribute to the choice of the FPN backbone. Disabling the shared linear transformations for \(\mathbf{Q}\) and \(\mathbf{K}\) in ET results in a collapse in performance, which is possibly due to the sparse supervision makes direct alignment challenging and increases redundancy.

\begin{figure}[t]
    \centering
    \includegraphics[width=0.49\textwidth]{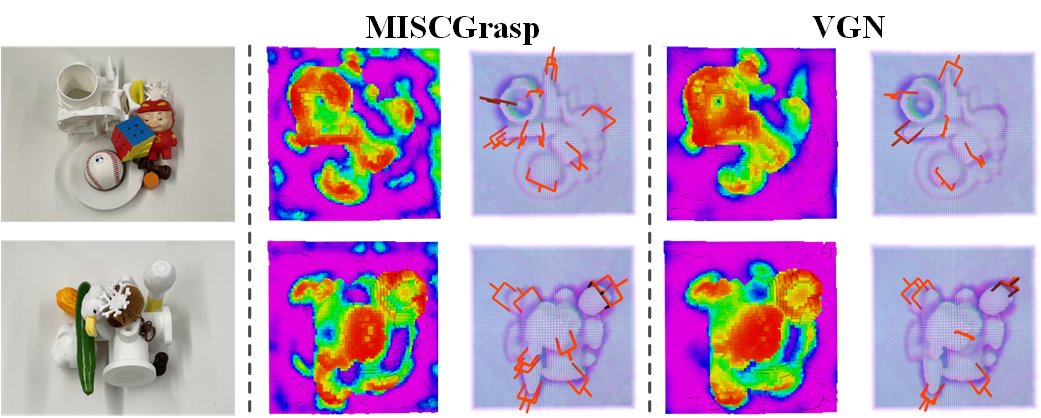}
    \vspace{-0.5cm}
    \caption{\textbf{Comparison of Visualization Results in Physical Multi-Object Experiments between MISCGrasp and VGN.} Our physical experiment scenes are notably more complex than those in previous works.}
    \label{fig6}
\end{figure}

\begin{table}[t]
    \centering
        \vspace{-0.2cm}
    \caption{RESULTS OF \textbf{SINGLE-OBJECT} EXPERIMENTS IN THE \textbf{Physical World}}
    \label{tab3}
    \small
    \resizebox{0.4\textwidth}{!}{  
    \begin{tabular}{l|ccc}
        \hline
        Method & SR (\%) & DR (\%) & Failure Cases \\
        \hline
        MISCGrasp & \cellcolor{blue!20}\textbf{80.0} & \cellcolor{blue!20}\textbf{80.0} & \cellcolor{blue!20}C3, D3, E3 \\
        \hline
        VGN & 60.0 & 60.0 & A3, B3, C3, D3, E2, E3 \\
        \hline
    \end{tabular}
    }
    \vspace{-0.5cm}
\end{table}

\begin{table}[t]
    \centering
        \vspace{-0.2cm}
    \caption{RESULTS OF \textbf{MULTI-OBJECT} EXPERIMENTS IN THE \textbf{Physical World}}
    \label{tab4}
    \small
    \resizebox{0.35\textwidth}{!}{  
    \begin{tabular}{l|cc}
        \hline
        \multirow{2}{*}{Method} & \multicolumn{2}{c}{Pile} \\
        \cline{2-3}
        & SR (\%) & DR (\%) \\
        \hline
        MISCGrasp & \cellcolor{blue!20}\textbf{90.3} (186 / 206) & \cellcolor{blue!20}\textbf{95.0} (190 / 200) \\
        \hline
        VGN & 65.7 (71 / 108) & 37.0 (74 / 200) \\
        \hline
    \end{tabular}
    }
    \vspace{-0.5cm}
\end{table}

\subsection{Physical Experiments}

\subsubsection{Setup}



Our experimental platform consists of a UR5 robotic arm equipped with a Robotiq 2-Finger 85 gripper, which performs grasping within a \(0.4 \times 0.4 \times 0.4 \ \text{m}^3\) workspace. An Intel RealSense D435 depth sensor is mounted on the robot’s wrist for perception.

We conduct both single-object and multi-object grasping in real-world experiments. For single-object experiment, $5$ EGAD objects are 3D-printed at sizes of \(3\) cm, \(6\) cm, and \(9\) cm to evaluate both power and pinch grasps.  Multi-object experiments use $9$ objects selected from EGAD, $14$ objects from Berkeley adversarial set, $30$ household items, and objects in the single-object experiment. The objects for real-world experiments are shown in Fig. \ref{fig5}. We use the same evaluation method as in simulation experiments. Multi-object experiments are conducted on pile scenarios which are rich in pinch grasps. Each test is repeated for $20$ rounds and there are $10$ objects in the scene.


\subsubsection{Results}


Tab. \ref{tab3} shows the results of single-object experiments. The failed grasp object IDs are also shown. VGN overly focuses on grasping on the edges of an object, which lead to some failures in execution. In contrast, MISCGrasp shows the ability of self-adaptive grasping through integration of both power and pinch grasping. 


Tab. \ref{tab4} shows that MISCGrasp outperforms VGN in real-world multi-object experiments. Our method declutters nearly all objects. It can be observed in Fig. \ref{fig6} that MISCGrasp generates more diverse grasp predictions. Please refer to the supplementary video for additional results and analysis. 

\section{CONCLUSIONS}


In this paper, we proposed MISCGrasp, a volumetric grasping method that leverages multi-scale feature integration and similarities among positive grasp samples to enhance grasping performance. Our method attends to fine-grained graspable local structures while simultaneously capturing global patterns, enabling self-adaptive grasping. Extensive experiments demonstrate its superiority, particularly in pile scenes rich in pinch grasps.


\bibliographystyle{IEEEtran}
\bibliography{IEEEabrv, IEEEexample, mybibfile}

\end{document}